%% file: main.tex
\documentclass{rvarxiv}

\input{preamble}
\usepackage[breaklinks,colorlinks,allcolors=rvblue,pdfcreator={},pdfproducer={}]{hyperref}
\pdftrailerid{}

\title{Seeing the Invisible: How Vision-Language Models\\Read, Predict and Trace Radio Fields}
\shorttitle{Seeing the Invisible: How Vision-Language Models Read, Predict and Trace Radio Fields}

\logos{%
  \rvlogo[9mm]{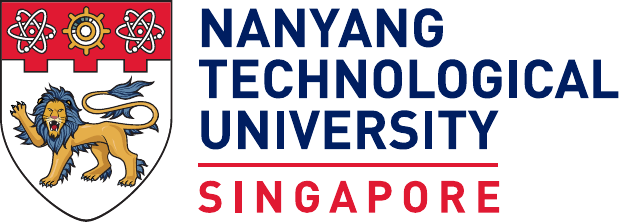}\hfill
  \rvlogo[9mm]{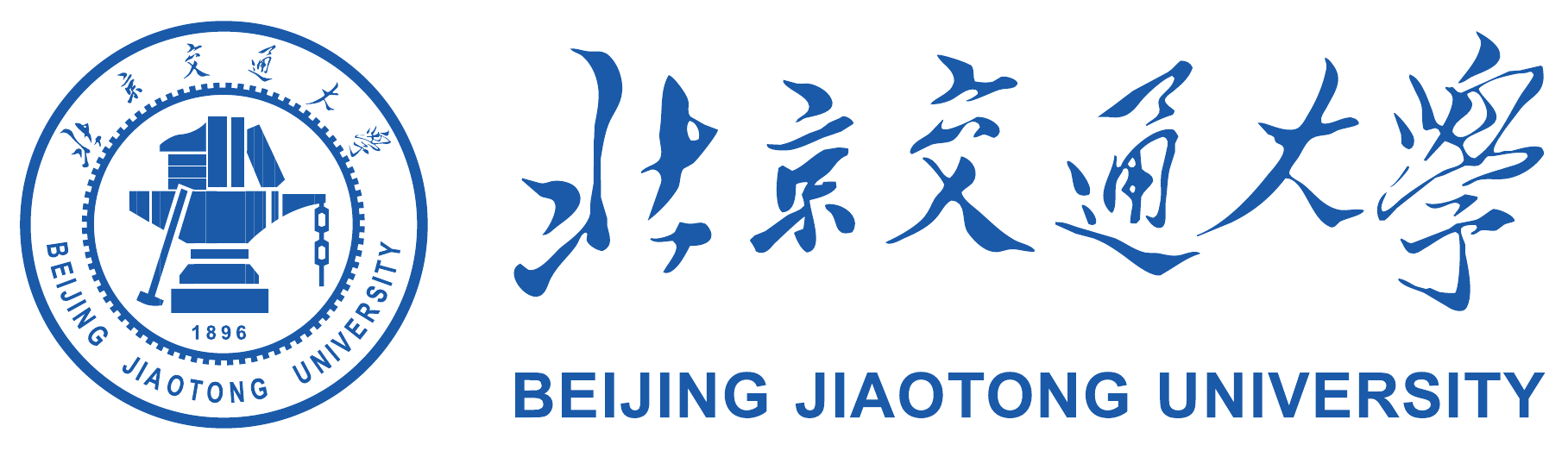}\hfill
  \rvlogo[9.5mm]{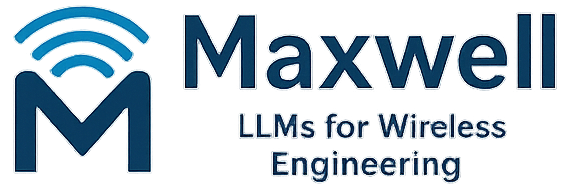}\hfill
  \rvlogo[8.5mm]{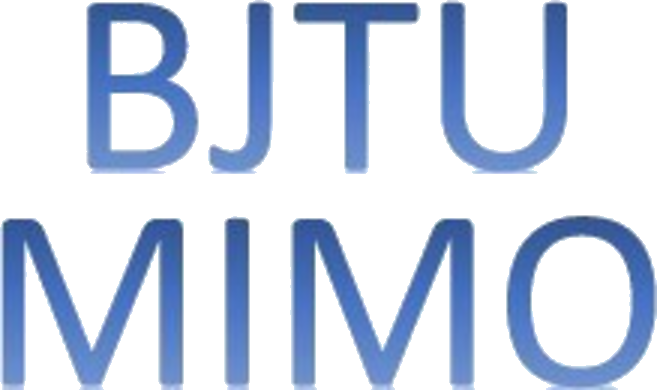}\hfill
  \rvlogo[9.5mm]{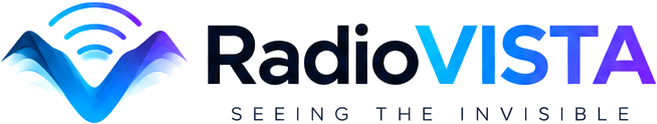}}

\author{%
  \textbf{Yiyang Zhu}\textsuperscript{1}, \textbf{Xin Li}\textsuperscript{1},
  \textbf{Guangyuan Liu}\textsuperscript{1}, \textbf{Ziheng Liu}\textsuperscript{2}\\
  \textbf{Jiayi Zhang}\textsuperscript{2}, \textbf{Chau Yuen}\textsuperscript{1}}
\affiliation{\textsuperscript{1}Nanyang Technological University,
  \textsuperscript{2}Beijing Jiaotong University}

\metadata{%
  \metaitem{Date}{November 18, 2026}%
  \metaitem{Project page}{\url{https://radiovista.github.io}}%
}

\input{sec/0_abstract}

\teaser{%
  \centering
  \includegraphics[width=\textwidth]{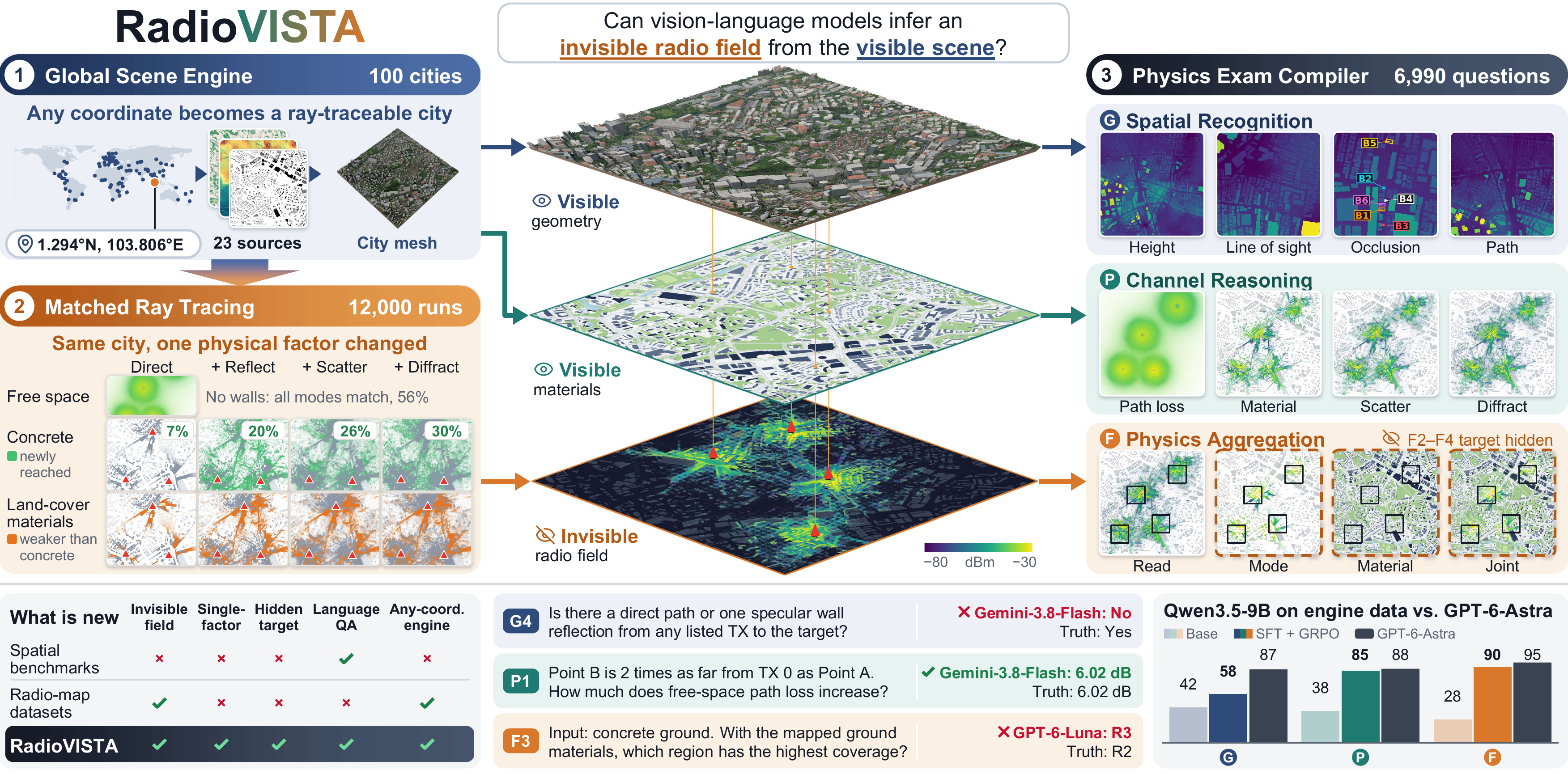}
  \caption{\textbf{Can vision-language models infer an invisible radio field from the visible scene?}
  Perceiving a city's geometry and materials, inferring the radio field they shape and predicting how it shifts under changing physics are central to invisible-field intelligence. Can VLMs ``see the invisible'' from visual cues alone, and where does their physical reasoning break down? We investigate this with \paper, which converts any coordinate into a ray-traceable city (1), perturbs one physical factor at a time (2), compiles the resulting fields into a physics exam with simulated ground truth (3) and probes what VLMs perceive and learn (bottom).}
  \label{fig:teaser}}

\begin{document}
\maketitle
\input{sec/1_intro}
\input{sec/2_related}
\input{sec/3_intelligence}
\input{sec/4_benchmark}
\input{sec/5_evaluation}
\input{sec/6_see}
\input{sec/7_learn}
\input{sec/8_conclusion}
\FloatBarrier
{
    \small
    \bibliographystyle{ieeenat_fullname}
    \bibliography{main}
}

\end{document}

%% file: preamble.tex
\usepackage{tikz}
\usetikzlibrary{positioning,arrows.meta,fit,calc,backgrounds,decorations.pathreplacing,shapes.geometric}
\usepackage{makecell}
\usepackage{colortbl}
\usepackage{array}
\usepackage{pifont}
\usepackage{placeins}
\usepackage{microtype}
\usepackage[scaled=0.92]{helvet}

\newcommand{\paper}{RadioVISTA\xspace}
\newcommand{\G}{\textsc{G}\xspace}
\newcommand{\Pt}{\textsc{P}\xspace}
\newcommand{\F}{\textsc{F}\xspace}
\newcommand{\cmark}{\ding{51}}
\newcommand{\xmark}{\ding{55}}
\definecolor{tierG}{HTML}{2F4B7F}
\definecolor{tierP}{HTML}{1E7B72}
\definecolor{tierF}{HTML}{D06A1F}
\definecolor{figdark}{HTML}{232B3A}
\definecolor{goodc}{HTML}{1E7B72}
\definecolor{badc}{HTML}{C0392B}
\definecolor{tabG}{HTML}{534AB7}
\definecolor{tabP}{HTML}{0F6E56}
\definecolor{tabF}{HTML}{993C1D}
\newcommand{\elev}[1]{\begin{tikzpicture}[baseline=-0.6ex,x=0.155cm]
\foreach \i in {1,...,11}{\ifnum\i>#1\relax\draw[black!30,line width=0.35pt] (\i,0) circle (0.052cm);\else\fill[tierF] (\i,0) circle (0.058cm);\fi}
\end{tikzpicture}}
\newcommand{\grpsep}{\arrayrulecolor{black!20}\midrule\arrayrulecolor{black}}
\newcommand{\bestA}[2]{{\color{#1}\bfseries #2}}
\newcommand{\bestK}[1]{{\bfseries #1}}
\newcommand{\bestU}[1]{\underline{#1}}
\newcommand{\bestO}[2]{{\color{#1}\bfseries\itshape #2}}
\newcommand{\pct}[1]{\begin{tikzpicture}[baseline=-0.6ex,x=0.155cm]
\pgfmathsetmacro{\pctv}{#1/10}
\foreach \i in {1,...,10}{
  \draw[black!30,line width=0.35pt] (\i,0) circle (0.052cm);
  \pgfmathsetmacro{\pctf}{max(0,min(1,\pctv-\i+1))}
  \ifdim\pctf pt>0pt
    \begin{scope}\clip (\i-0.058/0.155,-0.07) rectangle ++({\pctf*2*0.058/0.155},0.14);
    \fill[tierF] (\i,0) circle (0.058cm);\end{scope}
  \fi}
\end{tikzpicture}}

\newtcolorbox{takeawaybox}{enhanced,colback=tierG!6,colframe=tierG!6,boxrule=0pt,arc=3pt,
  left=5pt,right=5pt,top=2.5pt,bottom=2.5pt,boxsep=0pt,before skip=4pt,after skip=4pt,
  borderline west={2.2pt}{0pt}{tierG},fontupper=\itshape\small}
\newcommand{\takeaway}[1]{\begin{takeawaybox}#1\end{takeawaybox}}

%% file: sec/0_abstract.tex
\abstract{%
Operating in the physical world requires reasoning beyond what cameras directly capture. In a city, radio signals are invisible to the lens, yet their strength is governed by the buildings and streets an image does show. Given only these visible structures, can vision-language models (VLMs) infer the invisible radio field they shape? Probing true physical reasoning requires exact fields in real cities, alongside controlled interventions where a single physical factor changes. We build a label-free engine that turns any coordinate on Earth into a ray-traceable city, simulates matched single-factor counterfactuals, and computes every answer exactly. This yields \paper, a benchmark and training dataset for visual reasoning about invisible fields. Physically, a radio field is the sum of its propagation paths, yet across six continents we uncover a striking field-path dissociation: frontier VLMs accurately judge where a field is strong, but fail to tell whether a single direct or one-wall path exists. Open models fail even earlier at reading the field itself, and are systematically wrong, scoring below random guessing even when credited for any of four attempts. On unseen cities, fine-tuning brings a 9B open model close to frontier proprietary models on fields, yet neither supervised nor reinforcement fine-tuning improves path tracing. Supervising only final answers teaches models to approximate the field, yet leaves single-path geometry out of reach.
}

%% file: sec/1_intro.tex
\section{Introduction}
\label{sec:intro}

Many properties (e.g., depth, object shape, and surface material) of a scene are never directly recorded in an image, yet they can still be inferred from it. A city's radio field is likewise shaped by the building geometry and materials visible in the scene, yet it leaves no trace in any pixel. The strongest signal may reach a street around a corner, traveling along a reflected path no camera captures. Inferring this field requires reasoning beyond the image, from the visible structure it shows to the invisible physics at play.

We call this capability \emph{invisible-field intelligence}, inferring a physical field never shown in an image from the visible geometry and materials of the scene. It extends visual-spatial~\cite{yang2025thinking} and physical~\cite{bear2021physion,chow2025physbench} understanding from what is in view to what the view determines. While vision-language models (VLMs) reason fluently about the space they see~\cite{yang2025thinking,chen2024spatialvlm,tong2024cambrian}, whether they can infer a field beyond the image remains untested. Answering this matters for sixth-generation networks built around urban digital twins~\cite{saad2020vision,alkhateeb2023digitaltwin} and for language-model agents tasked with planning them~\cite{maatouk2023teleqna,hussain2025radiosim,liu2026malam4com}.

In a radio field, physics dictates a clear causal chain where geometry determines each path, materials govern signal attenuation along it, and the field sums thousands of such paths. Inferring this field therefore requires geometric competence to trace paths, propagation competence to judge how physical changes alter signals, and their integration across an entire region (\cref{sec:intelligence}). In this view, the path comes first, and the field is built from it.

Testing whether VLMs follow this physical account requires both the true field of a real city and its counterfactual after a single cause changes. Field measurements cannot switch off a city's reflections, but a simulator can. We therefore build an engine that turns any coordinate on Earth into a ray-traceable city, fusing 23 open geospatial sources into one watertight scene and extending Sionna RT with ground materials, tree canopy, and diffraction. It traces each city as a controlled experiment across twelve matched fields, where pairs isolate a single physical factor, and computes every answer exactly. This engine yields \paper (\cref{fig:teaser}), a benchmark for visual reasoning about invisible radio fields alongside scalable training data from the same pipeline (\cref{sec:bench}).

Contrary to this physical order, the strongest VLMs grasp the field without tracing the path. Across 19 evaluated models, the top two identify the best-covered urban region in roughly nine out of ten cases. Yet when asked whether a single direct or one-wall path exists, their accuracy drops to 76.5\% and 61.8\% against a 50\% random baseline. We term this gap \emph{field--path dissociation}, where models judge the whole without resolving its parts (\cref{sec:eval,sec:readtrace}). Open models fail even earlier at reading the field itself, staying systematically wrong and below random guessing despite sampling and feedback (\cref{sec:retry}).

Exact labels trace this dissociation to its source. Changing propagation physics rarely shifts the best-covered region, so models can largely read the field from the provided reference, whereas a path is never shown and must be traced. Training confirms this mechanism. Fine-tuning a 9B open model on unseen cities brings it close to the strongest proprietary model in channel reasoning and physics aggregation, yet it answers No on every path question. Outcome supervision thus conveys the field but not single-path geometry, which the reflection points computed by our engine can supply. In summary, our contributions are as follows:
\begin{itemize}
\item \textbf{Invisible-field intelligence and \paper}. We introduce invisible-field intelligence as a new axis of VLM physical reasoning, defined as inferring an unseen physical field from visible scene geometry and materials. For radio fields, we decompose this ability into geometric competence, propagation competence, and their integration, evaluated through spatial recognition, channel reasoning, and physics aggregation. To measure it, we build \paper with a label-free engine that turns any Earth coordinate into a ray-traceable 3D city, perturbs one physical factor at a time, and computes exact answers.
\item \textbf{Field--path dissociation in VLMs}. Evaluating 19 VLMs uncovers a sharp disconnect. The two strongest models localize peak field intensity under hidden interventions at up to 94.8\% accuracy, yet determine whether a direct or one-wall path exists in only 76.5\% and 61.8\% of cases. Open models reach at most 47.4\% overall and fall below random guessing even under Pass@4.
\item \textbf{Learnable fields, unlearned paths}. Post-training a 9B VLM on 68 cities boosts physics aggregation on 17 unseen cities from 27.9\% to 88.7--89.7\%, surpassing GPT-6-Astra on two channel-reasoning tasks. Yet neither supervised fine-tuning (SFT) nor group relative policy optimization (GRPO)~\cite{shao2024deepseekmath} improves path reachability (G4), pinpointing the boundary of outcome supervision.
\end{itemize}

%% file: sec/2_related.tex
\section{Related Work}
\label{sec:related}

\paragraph{VLMs with Spatial and Physical Awareness.}
Existing benchmarks test either visible space or few-object physics, rather than physics unfolding across a real city (\cref{tab:related}). Spatial benchmarks show that VLMs often misjudge relations, depth, and layout across images, videos, and maps~\cite{kamath2023whatsup,fu2024blink,tong2024cambrian,yang2025thinking,mukhopadhyay2024mapwise}, while synthetic question-answer pairs improve these judgments~\cite{chen2024spatialvlm,cheng2024spatialrgpt}. Yet every graded quantity remains in view. Physical benchmarks evaluate dynamics, counterfactuals, and material behavior~\cite{yi2020clevrer,bear2021physion,zheng2024contphy,chow2025physbench}, typically over a few objects in simple, mostly synthetic scenes. Whether VLMs can infer an unseen physical field governed by a real, cluttered city remains open.

\paragraph{Benchmarking Radio Fields.}
Radio datasets contain this unseen field, yet none can test VLMs on it because they fix the city or physics and lack language queries (\cref{tab:related}). Radio map estimation regresses ray-traced path loss~\cite{levie2021radiounet}, producing maps tied to training settings without revealing how geometry shapes propagation. Ray-traced datasets and digital twins link propagation to real geography~\cite{alkhateeb2019deepmimo,li2024geo2sigmap,testolina2024boston,hoydis2023sionna}, but most are limited to predefined scenarios, single cities, or fixed maps without customization. Geo2SigMap builds new scenes yet models materials only partially and offers no single-factor interventions. Wireless language models answer telecom queries, invoke simulators, or classify signals~\cite{maatouk2023teleqna,li2026wirelessmathbenchxl,hussain2025radiosim,han2026pred,liu2026scalinglaw}, ignoring the geometry and materials governing them. 

\begin{table}[t]
\centering
\caption{\textbf{\paper combines what prior work keeps apart.} Invisible field: the graded quantity is not visible in the input. Hidden target: the field after an intervention is never shown. A half-filled circle marks partial support.}
\label{tab:related}
\providecommand{\ymark}{\textcolor{tierP}{\cmark}}\providecommand{\nmark}{\textcolor{black!30}{\xmark}}%
\providecommand{\pmark}{\tikz[baseline=-0.6ex]{\fill[tierP] (0,0) -- (90:0.55ex) arc (90:270:0.55ex) -- cycle; \draw[tierP,line width=0.3pt] (0,0) circle (0.55ex);}}
\scriptsize
\setlength{\tabcolsep}{2.5pt}
\resizebox{\linewidth}{!}{%
\begin{tabular}{@{}lccccccc@{}}
\toprule
 & \makecell{Real 3D\\cities} & \makecell{Invisible\\field} & \makecell{Material\\aware} & \makecell{Single-factor\\intervention} & \makecell{Hidden\\target} & \makecell{Language\\QA} & \makecell{Any-coord.\\engine} \\
\midrule
VSI-Bench~\cite{yang2025thinking} & \nmark & \nmark & \nmark & \nmark & \nmark & \ymark & \nmark \\
SpatialVLM~\cite{chen2024spatialvlm} & \nmark & \nmark & \nmark & \nmark & \nmark & \ymark & \nmark \\
PhysBench~\cite{chow2025physbench} & \nmark & \nmark & \ymark & \nmark & \nmark & \ymark & \nmark \\
CLEVRER~\cite{yi2020clevrer} & \nmark & \nmark & \nmark & \ymark & \ymark & \ymark & \nmark \\
ContPhy~\cite{zheng2024contphy} & \nmark & \nmark & \ymark & \ymark & \ymark & \ymark & \nmark \\
TeleQnA~\cite{maatouk2023teleqna} & \nmark & \nmark & \nmark & \nmark & \nmark & \ymark & \nmark \\
RadioMapSeer~\cite{levie2021radiounet} & \ymark & \ymark & \nmark & \nmark & \nmark & \nmark & \nmark \\
Geo2SigMap~\cite{li2024geo2sigmap} & \ymark & \ymark & \pmark & \nmark & \nmark & \nmark & \ymark \\
\midrule
\rowcolor{figdark}\textcolor{white}{\textbf{\paper}} & \textcolor{white}{\cmark} & \textcolor{white}{\cmark} & \textcolor{white}{\cmark} & \textcolor{white}{\cmark} & \textcolor{white}{\cmark} & \textcolor{white}{\cmark} & \textcolor{white}{\cmark} \\
\bottomrule
\end{tabular}}
\end{table}

\begin{figure*}[t]
\centering
\includegraphics[width=\linewidth]{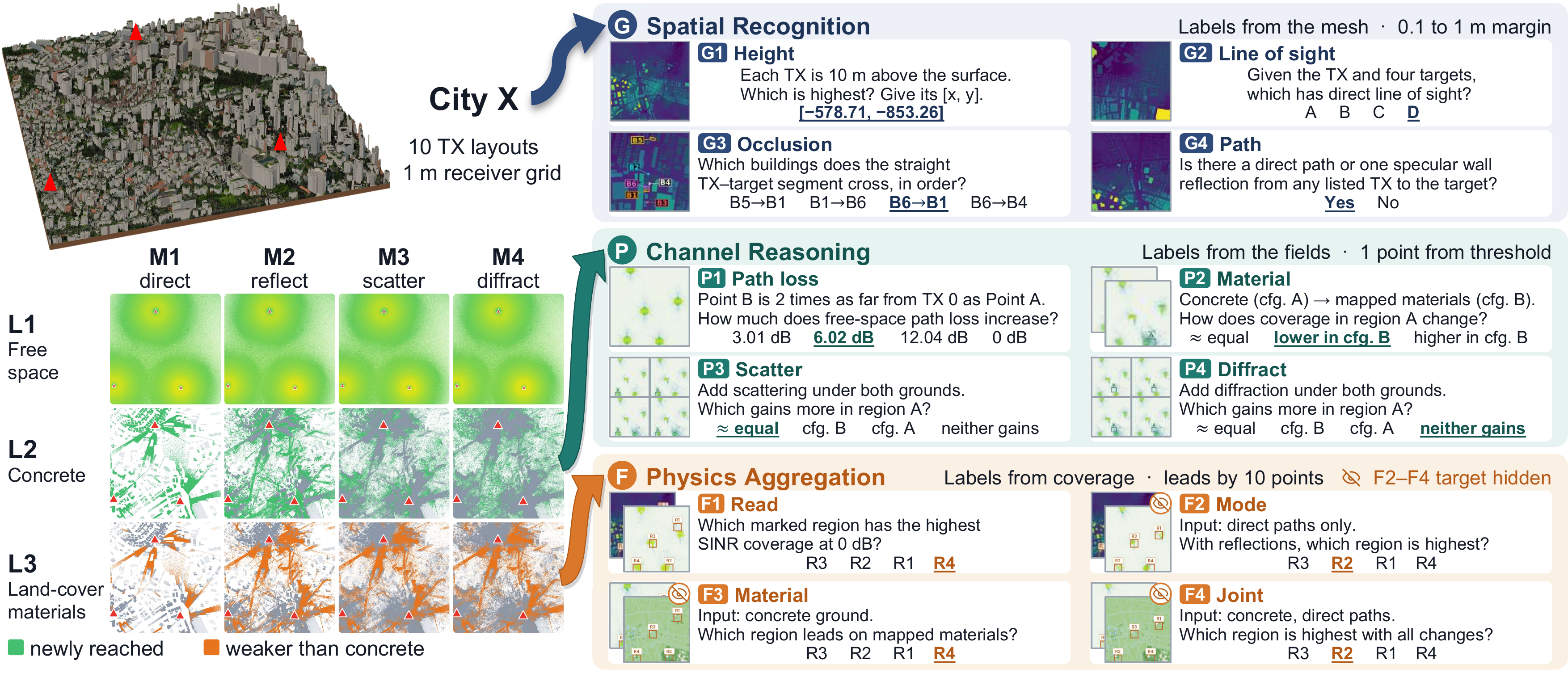}
\caption{\textbf{One engine, any city, twelve tasks.} Each city is ray-traced under three environments ($L_1$--$L_3$) and four propagation modes ($M_1$--$M_4$), and the matched fields yield twelve tasks of three types.}
\label{fig:taxonomy}
\end{figure*}

%% file: sec/3_intelligence.tex
\section{Invisible-Field Intelligence}
\label{sec:intelligence}

Visual-spatial intelligence concerns the space a model can see~\cite{yang2025thinking}. Invisible-field intelligence goes further, targeting a physical quantity absent from the input yet governed by visible geometry and materials. We ask a VLM to observe a real city and infer the radio field unfolding across it.

\paragraph{Two Competences and Their Combination.}
Consider a rooftop transmitter and a street behind a block of buildings. Geometry determines whether the signal reaches the street directly or off a single wall. The \emph{geometric} competence answers this question by tracing a path and judging whether it clears buildings or reflects off a wall. Propagation physics, in turn, dictates how the signal responds to physical changes such as longer distance, different ground materials, or added propagation modes, which the \emph{propagation} competence evaluates. The field itself combines both, summing thousands of reflected, scattered, and diffracted paths. Judging where the field is strong requires both competences together. 

\paragraph{Three Task Types.}
\paper evaluates these two competences and their combination across three task types, totaling twelve tasks (\cref{fig:taxonomy}). \emph{Spatial recognition} (\G) targets geometric competence, where the path-reachability task G4 directly asks whether a direct or one-wall path exists. \emph{Channel reasoning} (\Pt) evaluates propagation competence, while \emph{physics aggregation} (\F) tests their combination by ranking four regions by coverage, defined as the share of cells exceeding a signal threshold. 

\paragraph{The Order Physics Prescribes.}
Physics imposes a natural order on these abilities. A path must exist before it can carry energy, and the field sums all such paths. A model mirroring this causality would trace paths first, then read and predict fields. \paper tests whether VLMs follow this order.

%% file: sec/4_benchmark.tex
\section{\paper}
\label{sec:bench}

\subsection{Overview}
\label{sec:overview}

\begin{figure}[t]
\centering
\includegraphics[width=\linewidth]{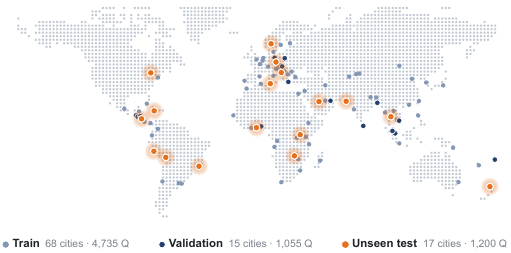}
\caption{\textbf{One hundred cities on six continents, none of the test cities seen in training.} The 100 released cities by split.}
\label{fig:stats}
\end{figure}

Testing this causal order requires both a city's true field and its counterfactual after a single cause changes. \paper obtains both through automated generation, taking an Earth coordinate as input and outputting exactly graded questions. Our engine solves four problems in sequence. It converts open geospatial data into a unified ray-traceable scene, simulates controlled runs to isolate individual physical factors, derives exact answers from the fields, and filters out ambiguous cases. Any new city or physical setting yields a fresh test suite without human annotation. Our current release spans 100 cities across six continents, with test cities strictly held out from training (\cref{fig:stats}).

\subsection{Benchmark Construction}
\label{sec:construction}

\paragraph{Scene Acquisition and Meshing.}
A ray tracer requires a single closed mesh with assigned surface materials, whereas open geospatial data arrives as separate layers for buildings, terrain, and land cover with varying coverage and resolution. Through 23 source adapters, our engine parses these inputs into a unified representation containing building footprints and heights, terrain, roads, and land-cover materials mapped to International Telecommunication Union Radiocommunication Sector (ITU-R) standards~\cite{itu2040,itu527}. It then welds buildings and terrain into a watertight mesh where every building edge is shared by two faces, enabling the solver to recognize edges for diffraction. By default, the engine combines GlobalBuildingAtlas~\cite{zhu2025globalbuildingatlas}, FABDEM~\cite{hawker2022fabdem}, ESA WorldCover~\cite{zanaga2022worldcover}, and OpenStreetMap~\cite{haklay2008osm} to construct scenes for any coordinate on Earth.

\paragraph{Matched Ray Tracing.}
A field summing all propagation mechanisms cannot reveal individual contributions. We therefore simulate each scene as a factorial experiment that varies one physical factor while holding all others fixed. Extending Sionna RT~\cite{hoydis2023sionna} with ITU-R ground materials, canopy absorption, and diffraction across every receiver cell, our solver simulates three environments covering free space ($L_1$), concrete geometry with terrain ($L_2$), and full geometry with land-cover materials and tree canopy ($L_3$). Each is traced under four cumulative propagation modes, from direct paths ($M_1$) to added reflection ($M_2$), scattering ($M_3$), and diffraction ($M_4$). Comparing $L_2M_1$ with $L_2M_2$, for instance, isolates reflections. Each scene includes ten transmitter layouts, yielding twelve matched fields on a 1\,m receiver grid under user-configurable frequencies, layouts, modes, and materials. Against the Friis law in free space, our solver agrees within 3\,dB on 98\% of 3,000 distance rings, with a median offset of $+1.3$\,dB.

\begin{table*}[t]
\centering
\caption{\textbf{Evaluation results on \paper.} Zero-shot accuracy (\%) on the 17 test cities. Bold and underline mark the best baseline and the best open model, and bold italics mark fine-tuned entries at or above the best baseline. Retry reports Pass@4 and correction rate.}
\label{tab:main}
\scriptsize
\setlength{\tabcolsep}{3.1pt}
\renewcommand{\arraystretch}{0.97}
\resizebox{\textwidth}{!}{%
\begin{tabular}{@{}l cccc>{\columncolor{tabG!7}}c cccc>{\columncolor{tabP!7}}c cccc>{\columncolor{tabF!7}}c c cc@{}}
\toprule
 & \multicolumn{5}{c}{\textcolor{tabG}{\textbf{Spatial recognition (G)}}} & \multicolumn{5}{c}{\textcolor{tabP}{\textbf{Channel reasoning (P)}}} & \multicolumn{5}{c}{\textcolor{tabF}{\textbf{Physics aggregation (F)}}} & \multicolumn{1}{c}{} & \multicolumn{2}{c}{\textbf{Retry}} \\
\cmidrule(lr){2-6}\cmidrule(lr){7-11}\cmidrule(lr){12-16}\cmidrule(lr){18-19}
Model & G1 & G2 & G3 & G4 & \G & P1 & P2 & P3 & P4 & \Pt & F1 & F2 & F3 & F4 & \F & All & P@4 & Corr. \\
\midrule
Random & 20.0 & 25.0 & 25.0 & 50.0 & 30.0 & 25.0 & 33.3 & 25.0 & 25.0 & 27.1 & 25.0 & 25.0 & 25.0 & 25.0 & 25.0 & 27.4 & 70.7 & 34.1 \\
\midrule
\multicolumn{19}{@{}l}{\textit{Proprietary models}} \\
GPT-6-Astra & \bestA{tabG}{78.2} & \bestA{tabG}{92.2} & \bestA{tabG}{100.0} & \bestA{tabG}{76.5} & \bestA{tabG}{86.7} & \bestA{tabP}{100.0} & \bestA{tabP}{93.1} & 71.6 & 85.3 & 87.5 & \bestA{tabF}{95.1} & 90.2 & \bestA{tabF}{98.0} & \bestA{tabF}{96.1} & \bestA{tabF}{94.9} & \bestK{89.7} & 92.8 & \bestK{90.6} \\
GPT-6.1-Sol & 73.1 & 91.2 & 97.1 & 61.8 & 80.8 & \bestA{tabP}{100.0} & 92.2 & \bestA{tabP}{76.5} & \bestA{tabP}{87.3} & \bestA{tabP}{89.0} & 89.2 & \bestA{tabF}{92.2} & 97.1 & 95.1 & 93.4 & 87.7 & \bestK{93.2} & 84.6 \\
GPT-6-Luna & 57.7 & 61.8 & 88.2 & 47.1 & 63.7 & 98.0 & 76.5 & 49.0 & 42.2 & 66.4 & 31.4 & 61.8 & 80.4 & 64.7 & 59.6 & 63.2 & 79.8 & 62.6 \\
Claude-Opus-5.5 & 69.2 & 70.6 & 94.1 & 64.7 & 74.7 & \bestA{tabP}{100.0} & 85.3 & 65.7 & 54.9 & 76.5 & 81.4 & 81.4 & 92.2 & 87.3 & 85.5 & 78.9 & 87.5 & 69.3 \\
Gemini-3.8-Flash & 62.8 & 57.8 & 92.2 & 61.8 & 68.6 & \bestA{tabP}{100.0} & 89.2 & 51.0 & 45.1 & 71.3 & 85.3 & 80.4 & 96.1 & 81.4 & 85.8 & 75.3 & 84.5 & 70.3 \\
\textbf{Proprietary mean} & 68.2 & 74.7 & 94.3 & 62.4 & 74.9 & 99.6 & 87.3 & 62.7 & 62.9 & 78.1 & 76.5 & 81.2 & 92.7 & 84.9 & 83.8 & 79.0 & 87.6 & 75.5 \\
\midrule
\multicolumn{19}{@{}l}{\textit{Open models}} \\
Qwen3.5-9B & 29.5 & 28.4 & 59.8 & 51.0 & 42.2 & 57.8 & 38.2 & 28.4 & 26.5 & 37.7 & 27.5 & 32.4 & 26.5 & 25.5 & 27.9 & 36.0 & 56.4 & 25.4 \\
Qwen3.8-27B & \bestU{38.5} & 33.3 & 58.8 & 52.9 & 45.9 & 72.5 & \bestU{71.6} & 41.2 & 27.5 & 53.2 & 40.2 & 27.5 & 32.4 & 31.4 & 32.8 & 44.0 & \bestU{68.3} & 50.3 \\
InternVL3.5-8B & 37.2 & 24.5 & 49.0 & 50.0 & 40.2 & 43.1 & 29.4 & 13.7 & 20.6 & 26.7 & 34.3 & 24.5 & 19.6 & 25.5 & 26.0 & 31.0 & 48.6 & 27.0 \\
InternVL3.5-38B & 32.1 & 40.2 & 28.4 & 52.9 & 38.4 & 70.6 & 26.5 & 25.5 & 25.5 & 37.0 & 25.5 & 22.5 & 19.6 & 22.5 & 22.5 & 32.7 & 52.5 & \bestU{50.4} \\
MiniCPM-V-4.7-1B & 35.9 & 28.4 & 51.0 & 52.9 & 42.1 & 11.8 & 44.1 & 26.5 & 24.5 & 26.7 & 21.6 & 23.5 & 28.4 & 24.5 & 24.5 & 31.1 & 44.9 & 27.7 \\
MiniCPM-V-4.7-35B & 30.8 & \bestU{42.2} & \bestU{78.4} & \bestU{57.8} & \bestU{52.3} & \bestU{96.1} & 41.2 & \bestU{54.9} & \bestU{32.4} & \bestU{56.1} & \bestU{47.1} & 21.6 & 40.2 & 26.5 & 33.8 & \bestU{47.4} & 65.1 & 39.8 \\
Molmo2-O-7B & 30.8 & 33.3 & 44.1 & 50.0 & 39.6 & 51.0 & 0.0 & 0.0 & 0.0 & 12.7 & 31.4 & \bestU{37.3} & \bestU{42.2} & 34.3 & \bestU{36.3} & 29.5 & 53.4 & 4.5 \\
Molmo2-8B & 29.5 & 32.4 & 74.5 & 45.1 & 45.4 & 42.2 & 11.8 & 0.0 & 1.0 & 13.7 & 28.4 & 35.3 & 39.2 & \bestU{36.3} & 34.8 & 31.3 & 55.9 & 43.0 \\
Gemma-4-31B & 35.9 & 37.3 & 55.9 & \bestU{57.8} & 46.7 & 82.4 & 56.9 & 38.2 & 19.6 & 49.3 & 25.5 & 35.3 & 21.6 & 29.4 & 27.9 & 41.3 & 48.9 & 37.4 \\
GLM-4.6V-Flash & 30.8 & 29.4 & 73.5 & 47.1 & 45.2 & 44.1 & 26.5 & 21.6 & 22.5 & 28.7 & 25.5 & 36.3 & 23.5 & 25.5 & 27.7 & 33.9 & 46.2 & 5.1 \\
Step3-VL-10B & 28.2 & 28.4 & 55.9 & 52.9 & 41.4 & 71.6 & 35.3 & 32.4 & \bestU{32.4} & 42.9 & 28.4 & 21.6 & 37.3 & 22.5 & 27.5 & 37.2 & 58.6 & 27.1 \\
ERNIE-4.5-VL-28B & 30.8 & 30.4 & 28.4 & 53.9 & 35.9 & 85.3 & 36.3 & 21.6 & 23.5 & 41.7 & 24.5 & 29.4 & 23.5 & 30.4 & 27.0 & 34.8 & 48.0 & 35.4 \\
Ministral-3-14B & 33.3 & 30.4 & 26.5 & 46.1 & 34.1 & 58.8 & 30.4 & 5.9 & 4.9 & 25.0 & 25.5 & 19.6 & 20.6 & 25.5 & 22.8 & 27.3 & 52.6 & 43.4 \\
Muse-Glimmer-30B & 24.4 & 15.7 & 22.5 & 47.1 & 27.4 & 23.5 & 26.5 & 28.4 & 24.5 & 25.7 & 24.5 & 19.6 & 25.5 & 25.5 & 23.8 & 25.6 & 61.6 & 1.2 \\
\textbf{Open mean} & 32.0 & 31.0 & 50.5 & 51.3 & 41.2 & 57.9 & 33.9 & 24.2 & 20.4 & 34.1 & 29.3 & 27.6 & 28.6 & 27.5 & 28.2 & 34.5 & 54.4 & 29.8 \\
\midrule
\multicolumn{19}{@{}l}{\textit{Ours}} \\
Qwen3.5-9B & 29.5 & 28.4 & 59.8 & 51.0 & 42.2 & 57.8 & 38.2 & 28.4 & 26.5 & 37.7 & 27.5 & 32.4 & 26.5 & 25.5 & 27.9 & 36.0 & 56.4 & 25.4 \\
\quad + SFT & 25.6 & 46.1 & \bestO{tabG}{100.0} & 52.9 & 56.2 & \bestO{tabP}{100.0} & \bestO{tabP}{94.1} & \bestO{tabP}{76.5} & 68.6 & 84.8 & 91.2 & 85.3 & 90.2 & 88.2 & 88.7 & 76.6 & 88.9 & 39.0 \\
\quad + SFT + GRPO & 29.5 & 48.0 & \bestO{tabG}{100.0} & 52.9 & 57.6 & \bestO{tabP}{100.0} & \bestO{tabP}{94.1} & 75.5 & 71.6 & 85.3 & 92.2 & 86.3 & 90.2 & 90.2 & 89.7 & 77.5 & 82.8 & 37.8 \\
\bottomrule
\end{tabular}
}
\end{table*}

\paragraph{Question Compilation.}
Every answer is computed directly from the mesh and matched fields without human annotation. Labels for \G derive from mesh intersection, using the image method to locate reflection points in G4, while labels for \Pt and \F derive from the simulated fields. Let $\rho_t(\mathbf{x})$ denote the power received at cell $\mathbf{x}$ from transmitter $t$, and let $S(\mathbf{x})=\max_{t} \rho_t(\mathbf{x})$ denote the serving power across all transmitters. Given the set of valid receiver cells $V$, the coverage of a region $R$ is
\begin{equation}
C_{\tau}(R) = \left|\{\mathbf{x}\in R\cap V : S(\mathbf{x})\geq\tau\}\right| / |R\cap V|,
\label{eq:coverage}
\end{equation}
where $\vert{}\cdot\vert{}$ denotes set cardinality and $\tau=-90$\,dBm (F1 instead thresholds the signal-to-interference-plus-noise ratio, SINR, at 0\,dB). In Tokyo, for example, an F3 question preserves the reference transmitters, regions, and propagation mode while changing only ground materials and tree canopy. The resulting coverages reach 14.4\%, 40.9\%, 44.8\%, and 66.7\%, so R4 wins by 21.9 percentage points.

\paragraph{Physics-Exact Quality Control.}
An exact answer yields a fair question only when it avoids details finer than the scene resolution. Our compiler therefore filters out boundary cases where minor shifts in geometry or field strength could flip the label, requiring every question to clear a task-specific margin. For instance, a G4 path must clear buildings by at least 0.1\,m, and the winner of an \F question must lead the runner-up by ten percentage points of coverage (\cref{eq:coverage}). Failing candidates are discarded, and layouts without stable candidates produce no questions for that task. Every surviving question thus has one unambiguous answer, governed by the physics of its city.

%% file: sec/5_evaluation.tex
\section{Evaluation on \paper}
\label{sec:eval}

\subsection{Evaluation Setup}
\label{sec:setup}

\paragraph{Benchmark Models.}
We evaluate 19 VLMs across 13 model families, spanning frontier proprietary systems and open models (publicly released weights) from 1B to 38B parameters in both dense and mixture-of-experts architectures. The proprietary systems include GPT-6-Astra, GPT-6.1-Sol, GPT-6-Luna~\cite{openai2026gpt6astra}, Claude-Opus-5.5~\cite{anthropic2026claudeopus55}, and Gemini-3.8-Flash~\cite{google2026gemini38flash}. The open models comprise Qwen3.5/3.8~\cite{qwen2026qwen35,qwen2026qwen38}, InternVL3.5~\cite{wang2025internvl35}, MiniCPM-V-4.7~\cite{yu2025minicpmv45}, Molmo2~\cite{clark2026molmo2}, Gemma-4~\cite{gemma2026gemma4}, GLM-4.6V~\cite{glmv2025glm45v}, Step3-VL~\cite{huang2026step3vl}, ERNIE-4.5-VL~\cite{baidu2025ernie45}, Ministral-3~\cite{liu2026ministral3}, and Muse-Glimmer~\cite{abtahizadeh2026museglimmer}.\footnote{Where a technical report is unavailable, we cite the closest preceding release in the same family.} All baseline evaluations are conducted zero-shot across the 1,200 questions from the 17 held-out test cities using identical images, prompts, and greedy decoding wherever supported.

\paragraph{Metric Design.}
Every task has a closed candidate set: five listed transmitter coordinates for G1, Yes or No for G4, and option letters elsewhere. We report accuracy, parsing the final answer of each response deterministically and counting missing selections as incorrect. Group scores average their four tasks, and overall accuracy averages all twelve.

\paragraph{Chance-Level Baselines.}
The random baseline is the accuracy of uniform guessing over each task's candidate set, from 20.0\% on G1 to 50.0\% on G4 and 27.4\% overall.

\subsection{Main Results}
\label{sec:results}

\paragraph{The Strongest Models Read Fields but Miss Paths.}
If VLMs inferred fields following physical causality, path tracing would be their first mastery (\cref{sec:intelligence}). Leading proprietary models show the opposite order. GPT-6-Astra and GPT-6.1-Sol achieve 89.7\% and 87.7\% overall (\cref{tab:main}), excelling on the most physics-intensive tasks by aggregating regional fields at 94.9\% and 93.4\%, while all five proprietary models apply the free-space law on P1 with at least 98.0\% accuracy. In contrast, G4 poses a much narrower binary question of whether a single direct or one-wall path exists, determined purely by visible geometry. Yet the five models average just 62.4\%, with GPT-6-Luna falling below a coin flip. \Cref{sec:readtrace} dissects this field--path dissociation.\takeaway{Recalling a propagation formula is solved, yet judging whether a signal clears a building remains open.}

\paragraph{Open Models Do Not Read the Field.}
Open models fail one step earlier, struggling to read the very fields that proprietary systems parse with ease. The strongest open model, MiniCPM-V-4.7-35B, reaches 47.4\% overall, and none exceeds 36.3\% on physics aggregation against a 25.0\% random baseline even when a field is provided. Scaling parameters does not close this gap either. Within InternVL3.5, growing from 8B to 38B improves channel reasoning from 26.7\% to 37.0\%, yet spatial recognition and physics aggregation dip from 40.2\% to 38.4\% and from 26.0\% to 22.5\%. This disparity persists across every test city, where the proprietary mean exceeds the open mean across all 17 cities by at least 37.9 percentage points. \Cref{sec:see,sec:retry} examine whether open models misread the field or default to a fixed preference regardless of it.

%% file: sec/6_see.tex
\section{How VLMs \emph{See} Invisible Fields}
\label{sec:see}

Two questions remain open, whether a model ranking regions of a hidden field reasons about the intervention or merely reads the reference, and whether reading a field entails tracing its paths. Exact labels resolve both by separating reference reading from intervention reasoning (\cref{sec:intervention}), and field reading from single-path tracing (\cref{sec:readtrace}).

\subsection{Probing via Matched Interventions}
\label{sec:intervention}

\begin{table}[t]
\centering
\caption{\textbf{Hidden targets separate reference copying from intervention tracking.} Accuracy (\%) on all 306 F2--F4 questions and the 295 with a kept winner, and correct counts on the other 11.}
\label{tab:copyref}
\footnotesize
\setlength{\tabcolsep}{4pt}
\renewcommand{\arraystretch}{1.15}
\begin{tabular*}{\linewidth}{@{\extracolsep{\fill}}l cc l@{}}
\toprule
 & \multicolumn{2}{c}{Accuracy (\%)} & \multicolumn{1}{c}{Changed winner} \\
\cmidrule(lr){2-3}\cmidrule(l){4-4}
Model & All & Kept & \multicolumn{1}{c}{correct of 11} \\
\midrule
\textit{Copy the reference} & 96.4 & 100.0 & \elev{0}\,~0 \\
\grpsep
GPT-6-Astra & \textbf{94.8} & \textbf{96.9} & \elev{4}\,~4 \\
GPT-6.1-Sol & \textbf{94.8} & 96.3 & \elev{6}\,~6 \\
GPT-6-Luna & 69.0 & 70.2 & \elev{4}\,~4 \\
Claude-Opus-5.5 & 86.9 & 87.8 & \elev{7}\,~\textbf{7} \\
Gemini-3.8-Flash & 85.9 & 87.5 & \elev{5}\,~5 \\
\grpsep
Open models (mean) & 27.9 & 28.2 & \elev{2}\,~2.1 \\
\grpsep
Qwen3.5-9B-SFT & 87.9 & 89.8 & \elev{4}\,~4 \\
Qwen3.5-9B-SFT-GRPO & 88.9 & 90.8 & \elev{4}\,~4 \\
\bottomrule
\end{tabular*}
\end{table}
A hidden-target question shows a reference field and asks which region wins after an intervention such as adding reflections. Answering it draws on two skills, reading the reference winner and judging whether the intervention shifts it. Let $p$ denote the fraction of questions where the winner changes, with $A_{\mathrm{keep}}$ and $A_{\mathrm{change}}$ denoting the accuracies on kept- and changed-winner subsets. Overall accuracy $A$ across all hidden-target questions then decomposes as
\begin{equation}
A = (1-p)\,A_{\mathrm{keep}} + p\,A_{\mathrm{change}}.
\label{eq:decomp}
\end{equation}
\looseness=-1 Measuring $p$ in \cref{eq:decomp} uncovers a strong physical regularity: interventions rarely shift the winner, which changes on only 11 of 306 test questions ($p=3.6\%$). Adding reflections, scattering, diffraction, or real materials reshapes coverage cell by cell yet seldom alters which region is best covered. Consequently, $A_{\mathrm{keep}}$ measures reference reading, whereas $A_{\mathrm{change}}$ isolates whether a model tracks the intervention.

\begin{figure}[t]
\centering
\resizebox{\linewidth}{!}{%
\begin{tikzpicture}[x=1cm,y=1cm,every node/.style={font=\sffamily\fontsize{7}{8.4}\selectfont}]
\def\w{4.01}
\def\g{0.08}
\node[inner sep=0,anchor=north west] at (0,0) {\includegraphics[width=\w cm]{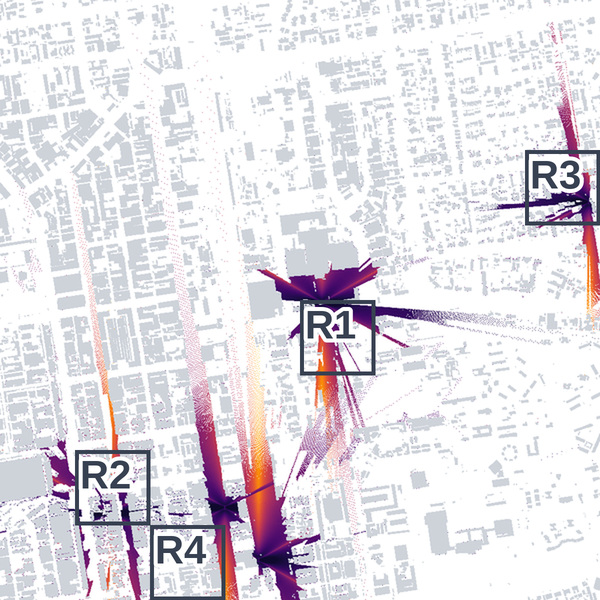}};
\draw[black,line width=0.25pt] (0,0) rectangle (\w,-\w);
\node[inner sep=0,anchor=north west] at (\w+\g,0) {\includegraphics[width=\w cm]{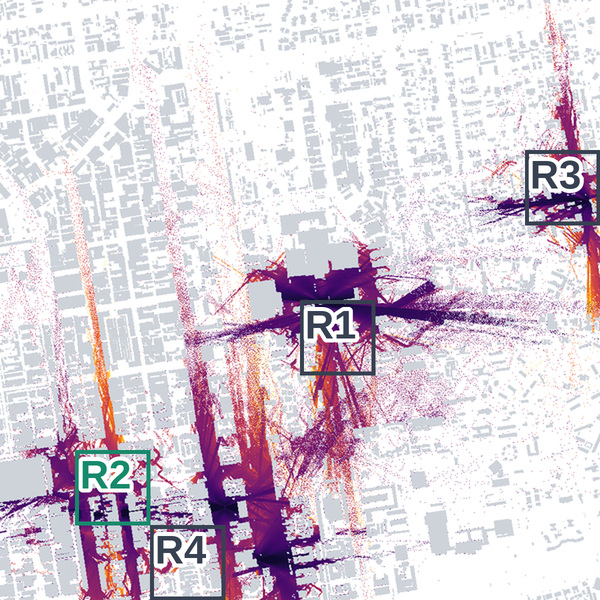}};
\draw[black,line width=0.25pt,dashed] (\w+\g,0) rectangle (2*\w+\g,-\w);
\node[anchor=north,inner sep=0] at (0.5*\w,-\w-0.1) {(a) F2 reference, $L_2M_1$};
\node[anchor=north,inner sep=0] at (1.5*\w+\g,-\w-0.1) {(b) hidden target, $L_2M_2$};
\node[anchor=north,inner sep=0,text=black!75,font=\sffamily\fontsize{6.4}{7.8}\selectfont] at (0.5*\w,-\w-0.46) {Highest coverage: R1};
\node[anchor=north,inner sep=0,text=black!75,font=\sffamily\fontsize{6.4}{7.8}\selectfont] at (1.5*\w+\g,-\w-0.46) {Highest coverage: R2};
\draw[black!25,line width=0.3pt] (0,-\w-0.84) -- (2*\w+\g,-\w-0.84);
\node[anchor=north,inner sep=0,align=center,text width=8.1cm,text=black!75,font=\sffamily\fontsize{6.4}{7.8}\selectfont] at (\w+0.5*\g,-\w-0.95) {%
Question: which region has the highest coverage in (b)?\\
GPT-6.1-Sol: R2 \textcolor{goodc}{\ding{51}}\qquad Claude-Opus-5.5: R1 \textcolor{badc}{\ding{55}}};
\end{tikzpicture}
}
\caption{\textbf{Wall reflections shift the winning region (F2).} Adding reflections moves the highest coverage from R1 in the visible reference (a) to R2 in the withheld target (b).}
\label{fig:lusaka}
\end{figure}

\paragraph{Kept and Changed Winners.}
Proprietary models handle the stable regime well and capture some intervention shifts (\cref{tab:copyref}). GPT-6-Astra and GPT-6.1-Sol exceed 96\% on kept-winner questions, and every proprietary model identifies the new winner on four to seven of the eleven changed-winner questions, where copying the reference winner scores zero. In Lusaka, for instance, GPT-6.1-Sol tracks the winner shifting from R1 to R2 once walls reflect, whereas Claude-Opus-5.5 retains R1 (\cref{fig:lusaka}(a,b)). In contrast, open models fail even in the stable regime, averaging 28.2\% on kept-winner questions near the 25\% random baseline and showing they fail to read the displayed field. Across bins of the reference winner's lead, proprietary accuracy rises from 73\% to 95\%, while the open mean stays below 30\%.
\takeaway{Proprietary models read the reference and track some interventions, whereas open models fail to read the provided field.}

\subsection{Reading Fields versus Tracing Paths}
\label{sec:readtrace}

\begin{figure}[t]
\centering
\includegraphics[width=\linewidth]{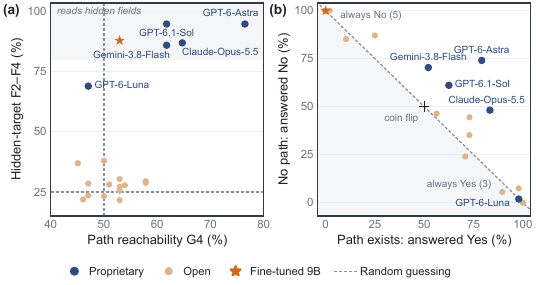}
\caption{\textbf{Field reading does not transfer to path tracing.} (a) Hidden-target (F2--F4) versus path reachability (G4) accuracy per model. (b) G4 accuracy on items with (48) and without (54) a valid path, where constant predictions fall in the corners.}
\label{fig:see}
\end{figure}

Reading a field draws on both competences simultaneously, yet kept-winner results show it can largely be read from a reference. In contrast, tracing a single path isolates geometric competence without any reference field, forcing models to infer the answer from geometry alone on \G questions. Here even the strongest models falter. GPT-6-Astra and GPT-6.1-Sol read and predict fields accurately in roughly nine out of ten cases, yet resolve whether a direct or one-wall path exists (G4) in only 76.5\% and 61.8\%. This gap defines \emph{field--path dissociation}, where models judge the aggregate outcome of thousands of paths while failing to verify a single one. Across all models, hidden-target accuracy spans from near random to near perfect, whereas G4 accuracy stays within 27 percentage points of the random baseline (\cref{fig:see}(a)).

\paragraph{Picking a Side Instead of Tracing a Path.}
Splitting G4 by its ground-truth label reveals why the task remains difficult (\cref{fig:see}(b)). GPT-6-Astra handles both classes well (79.2\% with a valid path and 74.1\% without), and GPT-6.1-Sol exceeds a coin flip on each, showing that they trace paths imperfectly. Most other models do not trace, collapsing onto a constant prediction of almost always No or Yes. Their near-50\% accuracy thus reflects label prevalence rather than geometric reasoning. Averaged across both classes, no open model beats a coin flip by more than 9 percentage points.
\takeaway{Reading a field and tracing a path are distinct abilities, where frontier VLMs master the former yet none traces paths reliably.}

%% file: sec/7_learn.tex
\section{How VLMs \emph{Learn} Invisible Fields}
\label{sec:learn}

Two failures are now clear, as open models fail to read fields and no model traces paths reliably. We test whether either is fixable via inference-time sampling and feedback (\cref{sec:retry}) or training on engine data (\cref{sec:tune}). Inference-time remedies resolve neither, whereas training a 9B open model fixes the former while leaving the latter untouched.

\subsection{Probing via Sampling and Feedback}
\label{sec:retry}

\begin{table}[t]
\centering
\caption{\textbf{Inference retries fail to rescue open models.} Pass@4 and correction rate (excluding G4), one dot per ten points. The best open model is taken per column, and random is uniform guessing.}
\label{tab:retry}
\footnotesize
\setlength{\tabcolsep}{3pt}
\renewcommand{\arraystretch}{1.15}
\begin{tabular*}{\linewidth}{@{\extracolsep{\fill}}l l l@{}}
\toprule
 & \multicolumn{1}{c}{Pass@4 (\%)} & \multicolumn{1}{c}{Correction rate (\%)} \\
 & \multicolumn{1}{c}{\scriptsize\color{black!50}random 70.7} & \multicolumn{1}{c}{\scriptsize\color{black!50}random 34.1} \\
\midrule
GPT-6-Astra & \pct{92.8}\,~92.8 & \pct{90.6}\,~\textbf{90.6} \\
GPT-6.1-Sol & \pct{93.2}\,~\textbf{93.2} & \pct{84.6}\,~84.6 \\
GPT-6-Luna & \pct{79.8}\,~79.8 & \pct{62.6}\,~62.6 \\
Claude-Opus-5.5 & \pct{87.5}\,~87.5 & \pct{69.3}\,~69.3 \\
Gemini-3.8-Flash & \pct{84.5}\,~84.5 & \pct{70.3}\,~70.3 \\
\textit{Proprietary mean} & \pct{87.6}\,~87.6 & \pct{75.5}\,~75.5 \\
\grpsep
\textit{Best open model} & \pct{68.3}\,~68.3 & \pct{50.4}\,~50.4 \\
\textit{Open mean} & \pct{54.4}\,~54.4 & \pct{29.8}\,~29.8 \\
\grpsep
Qwen3.5-9B-SFT & \pct{88.9}\,~88.9 & \pct{39.0}\,~39.0 \\
Qwen3.5-9B-SFT-GRPO & \pct{82.8}\,~82.8 & \pct{37.8}\,~37.8 \\
\bottomrule
\end{tabular*}
\end{table}

\paragraph{Sampling.}
If open models were merely uncertain, sampling multiple responses would lift them above random guessing, yet it does not. We draw four responses per question at temperature 0.7 and credit a question if any response is correct (Pass@4). Proprietary models average 87.6\%, whereas every open model remains below the 70.7\% expected from four uniform guesses and peaks at 68.3\% (\cref{tab:retry}), with Pass@$k$ curves falling below the random baseline from two samples onward.

\begin{figure}[t]
\centering
\includegraphics[width=\linewidth]{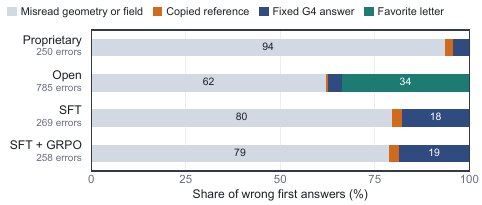}
\caption{\textbf{A third of open-model errors stem from a favorite letter.} Incorrect first responses by source for each model group, with errors per model out of 1,200 below.}
\label{fig:errors}
\end{figure}
\begin{figure}[t]
\centering
\includegraphics[width=0.92\linewidth]{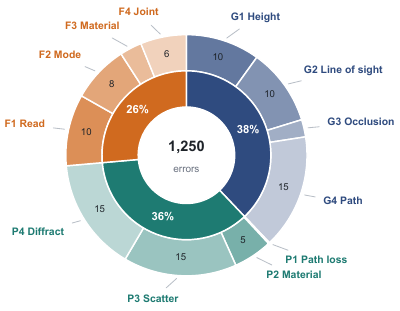}
\caption{\textbf{Paths and propagation dominate proprietary errors.} All 1,250 errors of the five proprietary models by capability group (inner ring) and task (outer ring, \% of total).}
\label{fig:errpie}
\end{figure}

\paragraph{Feedback.}
Telling a model its first answer is wrong helps only when it can perceive a valid alternative. Excluding binary G4, where feedback directly reveals the answer, proprietary models correct 75.5\% of their errors on a second attempt. Open models correct only 29.8\% on average, below the 34.1\% expected from uniform guessing among remaining options (\cref{tab:retry}, with per-model rates in \cref{tab:main}). Three open models correct fewer than one error in ten.

\paragraph{Why Errors Persist.}
Both remedies fail because open-model errors are systematic rather than noisy. Twelve of the fourteen open models assign over a third of their predictions to a single option letter, whereas ground-truth labels and proprietary models distribute evenly near a quarter each. Across four samples, open models repeat identical wrong answers on 25.2\% of questions, triple the rate of proprietary systems (8.1\%). Sampling therefore redraws the same choice, while feedback cannot redirect a model ignoring the visual scene. Scaling fails to eliminate this bias either, with accuracy correlating weakly with log parameter count ($r=0.38$).
\paragraph{Error Sources.}
Exact labels let us attribute every error without human annotation (\cref{fig:errors}). We assign each incorrect first response to the first source that applies: copying the reference winner when it shifts (F2--F4), a constant G4 label (at least 90\% of items), a favorite option letter (over a third of predictions), or a misreading of the geometry or field. Proprietary errors are almost entirely misreadings (93.5\%), and nearly half of them fall on G4, P3, and P4, where a single path or mechanism governs the outcome (\cref{fig:errpie}). For open models, in contrast, a favorite letter explains a third of the errors, which resolves the question raised in \cref{sec:eval}: they misread the field and fall back on fixed preferences. Training removes this letter bias but leaves a constant No on G4, which accounts for 17.8\% of the SFT model's errors.
\takeaway{Open-model errors are systematic, with a third following a fixed answer preference, and neither sampling nor feedback lifts these models above random guessing on average.}

\begin{figure}[t]
\centering
\includegraphics[width=0.83\linewidth]{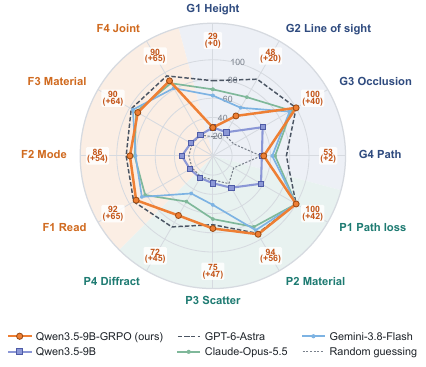}
\caption{\textbf{Training teaches the field, not the path.} Per-task accuracy of Qwen3.5-9B before and after fine-tuning, alongside three proprietary models. The outer ring gives the fine-tuned accuracy and its gain over the base model.}
\label{fig:finetune}
\end{figure}

\subsection{Learnable Fields, Unlearned Paths}
\label{sec:tune}

The engine generating the benchmark also synthesizes training data for cities strictly disjoint from the test set. Using low-rank adaptation (LoRA)~\cite{hu2022lora}, we fine-tune Qwen3.5-9B across 68 training cities, first via SFT and then via GRPO~\cite{shao2024deepseekmath} with an answer-correctness reward, while monitoring 15 validation cities.

\paragraph{Engine Data Teaches the Field.}
Across the 17 test cities, fine-tuning lifts our 9B model from 36.0\% to 77.5\% overall, surpassing Gemini-3.8-Flash and GPT-6-Luna while outperforming the strongest open model by 30.1 percentage points (\cref{tab:main}). Gains concentrate where open models struggled most, boosting physics aggregation from 27.9\% to 89.7\% near GPT-6-Astra's 94.9\%, and overtaking GPT-6-Astra on channel-reasoning tasks P2 and P3 (\cref{fig:finetune}). Trained solely on engine data, a compact 9B model thus rivals frontier proprietary systems at reading and predicting radio fields.

\paragraph{The Path Remains Out of Reach.}
Comparing the two training stages pinpoints the bottleneck. GRPO adds roughly one percentage point over SFT in each capability group (\cref{tab:main}), and both identify four of the eleven changed winners, matching GPT-6-Astra (\cref{tab:copyref}). On G4, however, both predict No on every item, yielding 52.9\% accuracy that merely matches the negative-class prevalence (\cref{fig:see}(b)). Fine-tuning thus reproduces the field--path dissociation of frontier proprietary systems within a 9B model. Supervision on the final answer, whether imitated or rewarded, teaches field aggregation without imparting single-path geometry.

%% file: sec/8_conclusion.tex
\section{Conclusion}
\label{sec:conclusion}

A city shapes its radio field path by path, leaving no trace in any image. To test whether VLMs can infer this invisible field, we built an engine that turns any coordinate on Earth into a ray-traceable city with exact answers, yielding \paper. Our findings uncover a striking field--path dissociation. Frontier VLMs judge where a field is strong yet often fail to verify a single path, and answer-only fine-tuning reproduces this same gap in a 9B open model on unseen cities. Fields can be read, whereas paths must be traced, and outcome supervision alone cannot teach tracing. Bridging this gap demands training models to reason through intermediate physics rather than merely fitting final answers.